\documentclass[conference]{IEEEtran}
\IEEEoverridecommandlockouts
\usepackage{cite}
\usepackage{citesort}
\usepackage{amsmath,amssymb,amsfonts}
\usepackage{algorithmic}
\usepackage{graphicx}
\usepackage{textcomp}
\usepackage{xcolor}
\usepackage{subfig}
\usepackage{multirow}

\def\BibTeX{{\rm B\kern-.05em{\sc i\kern-.025em b}\kern-.08em
    T\kern-.1667em\lower.7ex\hbox{E}\kern-.125emX}}

\usepackage[font=small,labelfont=bf,tableposition=top]{caption}

\newcommand{\ie}{\textit{i.e.}}

\newcommand{\etal}{\textit{et al.~}}

\title{MTRNet: A Generic Scene Text Eraser}

\begin{document}


\author{Osman Tursun, Rui Zeng, Simon Denman, Sabesan Sivapalan, Sridha Sridharan, Clinton Fookes\\
\textit {SAIVT, Queensland University of Technology}\\
\it \{w.tuerxun, r5.zeng, s.denman, s.sabesan, s.sridharan, c.clinton\}@qut.edu.au}

\maketitle


\begin{abstract}
Text removal algorithms have been proposed for uni-lingual scripts with regular shapes and layouts. However, to the best of our knowledge, a generic text removal method which is able to remove all or user-specified text regions regardless of font, script, language or shape is not available. Developing such a generic text eraser for real scenes is a challenging task, since it inherits all the challenges of multi-lingual and curved text detection and inpainting. To fill this gap, we propose a mask-based text removal network (MTRNet). MTRNet is a conditional adversarial generative network (cGAN) with an auxiliary mask. The introduced auxiliary mask not only makes the cGAN a generic text eraser, but also enables stable training and early convergence on a challenging large-scale synthetic dataset, initially proposed for text detection in real scenes. What's more, MTRNet achieves state-of-the-art results on several real-world datasets including ICDAR 2013, ICDAR 2017 MLT, and CTW1500, without being explicitly trained on this data, outperforming previous state-of-the-art methods trained directly on these datasets.

\end{abstract}


%
%
\section{Introduction}
\label{sec:intro}

Text removal refers to the task of erasing text present in a scene by replacing it with new content, which is semantically plausible and has realistic texture details. The task has drawn the attention of the computer vision community due to its potential value for privacy protection \cite{nakamura2017scene,zhang2018ensnet}, as text removal approaches automatically remove private information such as license plate numbers, addresses, and names from images shared via social media. It is also helpful for other applications including image editing, image restoration, \cite{nakamura2017scene,ye2015text} and image retrieval \cite{tursun2018component}. 

\begin{figure}[t!]
	\centering
	\includegraphics[width=0.47\textwidth]{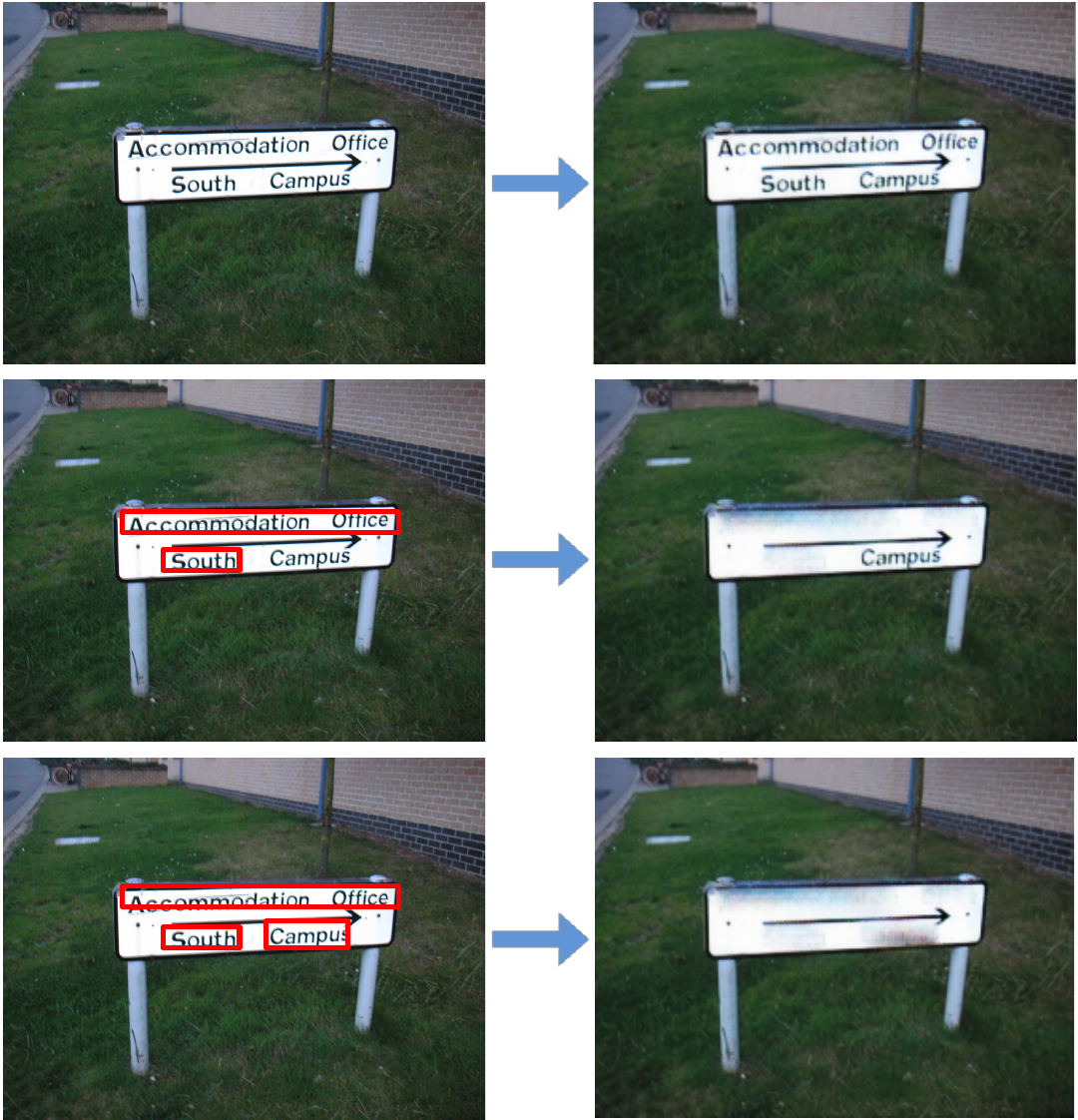}
	\caption{\textbf{Text removal using text masks:} the proposed mask-based text removal algorithm is capable of removing all or user specified text regions via inpainting them with visually plausible backgrounds. From top to down, in each example, the left image is input, while the right is the generated. Bounding boxes in red indicate the regions where text masks exist. \it{Note that the shown sample image is a real-world image, while the model is trained on a synthetic dataset}.}
	\label{fig:fig1}
\end{figure} 

An image transformation paradigm is applied for text removal in recent state-of-the-art studies \cite{nakamura2017scene,ye2015text} where a convolutional encoder-decoder model is trained in an end-to-end manner to generate text-free images for given input images. 
Compared to traditional methods, which are based on a two-step pipeline of: \textit{text detection} followed by \textit{inpainting}, they are able to capture the global structure of scenes and replace text with semantically plausible visual content. Moreover, they are more efficient as they skip the text detection process. Compared to hand-crafted approaches based on a two-step pipeline, they improve both qualitative and quantitative results by a large margin\cite{nakamura2017scene,ye2015text}.

However, current text removal methods based on end-to-end image transformation methods have the following disadvantages: (a). Those models do not support partial text removal. All text in a scene will be erased and as such these methods are not appropriate when partial text removal, which may be needed for image editing and retrieval, is required; (b). They have a limited generalization ability and fail in multi-lingual or distorted cases because the models are only valid for uni-lingual text data with limited shape and layout variability; and (c). Methods often exhibit late convergence and unstable training due to the challenging training datasets.



In this study, to overcome the aforementioned disadvantages of current state-of-the-art methods, a mask-based text removal network (MTRNet) is proposed. It is a conditional generative adversarial network (cGAN) \cite{isola2017image} with auxiliary masks. The introduced auxiliary masks prove to be very beneficial. Firstly, they not only relieve the cGAN of overcoming text detection challenges such as scene complexity, illumination and blurring, but also focus the network on the task of text inpainting. As a result, the cGAN achieves fast convergence on a challenging large-scale text detection dataset. Additionally, the masks can direct the cGAN to partially remove text as shown in Fig. \ref{fig:fig1}. Finally, they make the cGAN more generic. A cGAN trained with masks of a uni-lingual script \ie, English, with limited distortion, is effective for multi-lingual and distorted text without retraining or fine-tuning. 

Masks are also affordable from an annotation standpoint, and MTRNet only requires coarse text masks. Besides that, the cost of generating text masks for situations such as real-scene, multi-lingual and distorted text is decreasing; as text detection with advanced deep learning approaches achieves outstanding performance in challenging real-scene datasets \cite{karatzas2013icdar,yuliang2017detecting,nayef2017icdar2017}.

To summarize, our contribution is three-fold:
\begin{itemize}
	\item We propose a unified framework for text removal which combines the benefits of hand-crafted and deep learning methods. The experimental results on ICDAR 2013 and CTW1500 datasets demonstrates our method surpasses all previous state-of-the-art techniques.
	\item We also evaluate the proposed text removal algorithm on more challenging datasets such as the multi-script and curved datasets. The proposed method demonstrates state-of-the-art performance.
	\item We investigate the role of auxiliary masks for text removal. We show that masks significantly improve both qualitative and quantitative performance. Moreover, the training process is more stable and efficient with masks.
\end{itemize}

The rest of the paper is organized as follows: Sect. \ref{sec:lit} discusses related text removal works. The proposed method is introduced in detail in Sect. \ref{sec:met}. In Sect. \ref{sec:exp}, the experimental setup, datasets and results are provided. Finally, a brief conclusion is made in Sect. \ref{sec:con}.

%
%
\section{Related work}
\label{sec:lit}

Text removal methods are categorized into one-step \cite{nakamura2017scene} and two-step \cite{tursun2018component} based approaches. Here, related works on these two types of approach are discussed.

Two-step approaches remove text by inpainting text located with a text detection algorithm. Early two-step approaches \cite{khodadadi2012text,wagh2015text} are mainly based-on primitive hand-crafted text-detection and inpainting algorithms. A recent two-step approach with more accurate text localization is proposed by Tursun \etal \cite{tursun2018component}. They proposed pixel level text segmentation for text localization, however, inpainting is implemented by replacing text pixels with the most-frequently occurring neighborhood background color around the text region. Aforementioned two-step approaches lack of generality as they are designed for removing text in born-digital content such as subtitles and logos.



%
%
One-step approaches cast text removal as an image transformation task. Nakamura \etal \cite{nakamura2017scene} applied a deep skip connected encoder-decoder model for removing text in patches of images. Their method improved performance over a baseline text detection method. However, dividing images into many small patches causes a loss of context information. Without context, the proposed method fails to remove large text regions in a scene. Zhang \etal \cite{zhang2018ensnet} applied a conditional generative adversarial network (cGAN) for removing text for a whole image. Their network structure is very similar to the cGAN proposed by Isola \etal \cite{isola2017image} and boosts both qualitative and quantitative results of cGAN via multiple loss functions and a novel lateral connection.

Our proposed method is a novel two-step approach with a similar architecture to one-step approaches and casts text inpainting as image transformation problem.

%
%
\section{Method}
\label{sec:met}

\begin{figure*}[t!]
	\includegraphics[width=\textwidth]{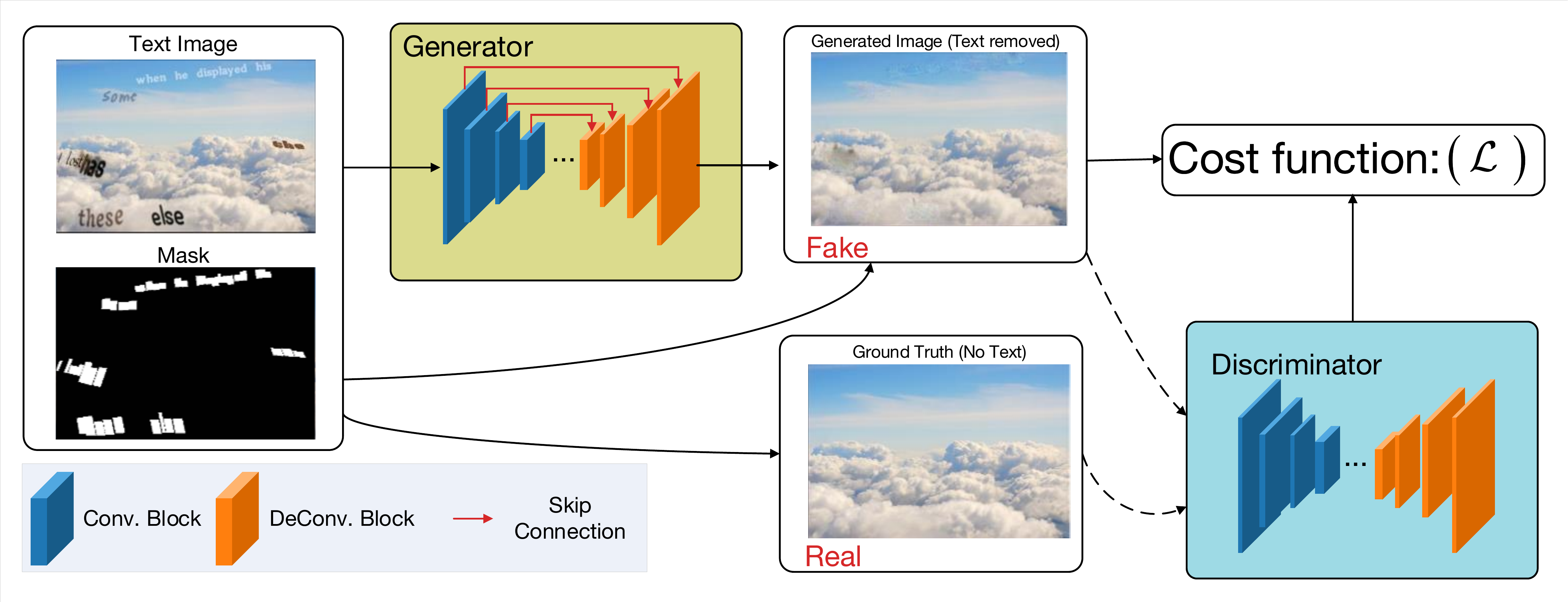}
	\caption{The overall structure of  MTRNet. It is composed of a generator and discriminator. The input to the generator is a concatenation of the input image and its text mask, and the input to the discriminator is a concatenation of the groundtruth/generator output and the input of the generator.}
	\label{fig:model}
\end{figure*}
\noindent
{\bfseries{Conditional generative adversarial networks (cGAN)}} constitute a framework which can be adapted for image-to-image translation \cite{isola2017image}. They are composed of a Generator ($G$) and a Discriminator ($D$). In an image-to-image transformation task, every pixel in an image is mapped from the original domain to the target domain. $G$ is trained to learn the mapping, while $D$ is used to evaluate the performance of $G$. The cGAN is trained with pairs of corresponding images ${(x, y)}$, where $x, y$ belongs to source and target domains respectively. The following objective loss is applied for training,

\begin{equation}
G^*=arg\min_{G}\max_{D}\mathcal{L}_{cGAN}(G,D) + \lambda \mathcal{L}_{L1}(G), 
\end{equation}

where $\mathcal{L}_{cGAN}(G,D)$ is the adversarial loss given in Eq. \ref{eq:gan}, and $\mathcal{L}_{L1}$ is the L1 loss given in Eq. \ref{eq:l1}. Here, the L1 loss is guiding the model to produce outputs which have the same general structure as the ground truth, while the adversarial loss teaches the model to approximate the distribution of the high-frequency signals that appear in the ground truth,

\begin{multline}
\mathcal{L}_{cGAN}= \min_{G}\max_{D}\mathbb{E}_{x,y \sim p_{data}(x,y)}[logD(x,y)]+ \\
\mathbb{E}_{x \sim p_{data}(x)}[log(1 - D(x, G(x))],
\label{eq:gan}
\end{multline}

\begin{equation}
\mathcal{L}_{L1}(G)=\mathbb{E}_{x,y\sim p_{data(x,y)}}[\left \|  y-G(x) \right \|_1].
\label{eq:l1}
\end{equation}

The cGAN provides an intuitively appealing solution for text removal as the latter is also an image transformation problem. However, due to the fact that text is sparse in real-life scenes, the source and target domain are very similar. The similarity between source and target domain is one of the reasons as to why training cGANs with real scene datasets is challenging. What's more, even a well-trained cGAN has a limited applicable scope.

In this study, the text mask of the input is employed to solve the aforementioned disadvantages of cGANs. The text mask of the input is sent to both G and D as an extra channel of the input. In other words, we replace $x$ with $(x, f(x))$, where $f$ is a function for generating text masks for $x$, which can be a text detection algorithm or manual input. During training, it boosts the convergence of both $G$, and $D$, while at inference it endows a generic and partial text removal ability to the model. 

The overall structure of MTRNet is given in Fig. \ref{fig:model}. It is composed of a G and a D, which are introduced in detail in the following sections.
\\[1.5ex]
\noindent
{\bfseries{Generator:}} In this work, U-Net \cite{ronneberger2015u} is the backbone of G. U-Net is a skip-connected fully convolutional neural (FCN) network. It is composed of an encoder and a decoder, which share convolutional channels. Those shared convolutional channels share information from the encoder to the decoder, providing useful context information for generating realistic images. U-Net receives an input of size $256 \times 256 \times 4$, which is a concatenation of red, green and blue channels of an image and the corresponding text-mask. All images and their corresponding masks are resized to 256 square while retaining the original aspect ratio though zero padding. The input is normalized to the range [-1, 1]. 

The encoder is composed of seven convolutional layers with kernel size 4, stride step 2, and zero padding. The number of kernels are $64, 128, 256, 512, 512, 512, 512$. LeakyReLU and batch normalization are applied in all Convolutional layers.

The decoder is composed of seven deconvolution layers with stride step 2 and eight convolutional layers with stride step 1. Deconvolution layers are applied to enlarge the size of the input of the decoder to the size of original image. The output of each deconvolution layer is sent to a subsequent convolutional layer, which stacks the output of the corresponding convolutional layer in the encoder. All layers have kernels with size 4, and zero padding. As with the encoder, batch normalization and LeakyRelu are applied to all layers except the last convolutional layer where a tanh function is used to map the output to the range [-1, 1].
%
%
\\[1.5ex]
\noindent
{\bfseries{Discriminator:}}
The discriminator architecture follows PatchGAN \cite{isola2017image}, which is designed to penalize the structure of the input (real image or generator output) at the scale of patches. It classifies if each patch as real or fake. In this work, the patch size of D is set to $1 \times 1$. Although a big patch size is beneficial for efficiency, the discriminator fails to focus on the text removal task because of the sparsity of text in a scene. With pixel-level patches, losses at text masks are back propagated to the model sufficiently.

A FCN is used as the backbone for D. Its output size is set to the same size as the input to get pixel level patches. As shown in Fig. \ref{fig:model}, it is composed of an encoder and a decoder. The encoder has four convolutional layers, while the decoder has four deconvolutional layers and one convolutional layer. In the encoder, the number of kernels are $64, 128, 256, 512$, while in the decoder, the number of kernels are $512, 256, 128, 64, 1$. All convolutional and deconvolutional layers except the last convolutional layer have kernels with the size of $4$, and a step stride of 2. They are also activated by LeakyRelu, and batch normalized. However, no activation and normalization are applied to the last convolutional layer. Its stride size is set to 1. Note zero padding is applied to all layers.
\\[1.5ex]
\noindent
{\bfseries{Training:}}
We have implemented our model with and without the mask using TensorFlow. Adam is chosen as the optimizer and its momentum is set to 0.9. The initial learning rate is 0.0002, and decays by a factor 0.95 after every 2 epochs. The size of an epoch is 50,000 iterations. The batch size is set to 16. The model is trained for 6 epochs.
%
\section{Experiments}
\label{sec:exp}

Experiments are performed to answer the following questions: 1) Do auxiliary masks help a GAN converge faster? 2) Do masks assist a GAN in text removal? and 3) Does this simple model generalize to other related real-word tasks such as multilingual and curved texts when the model is trained on a synthetic dataset with English characters in a  conventional layout?

\subsection{Datasets and Evaluation Metrics}
\begin{figure}[!t]
	\centering
	\subfloat{
		\includegraphics[width=0.45\linewidth]{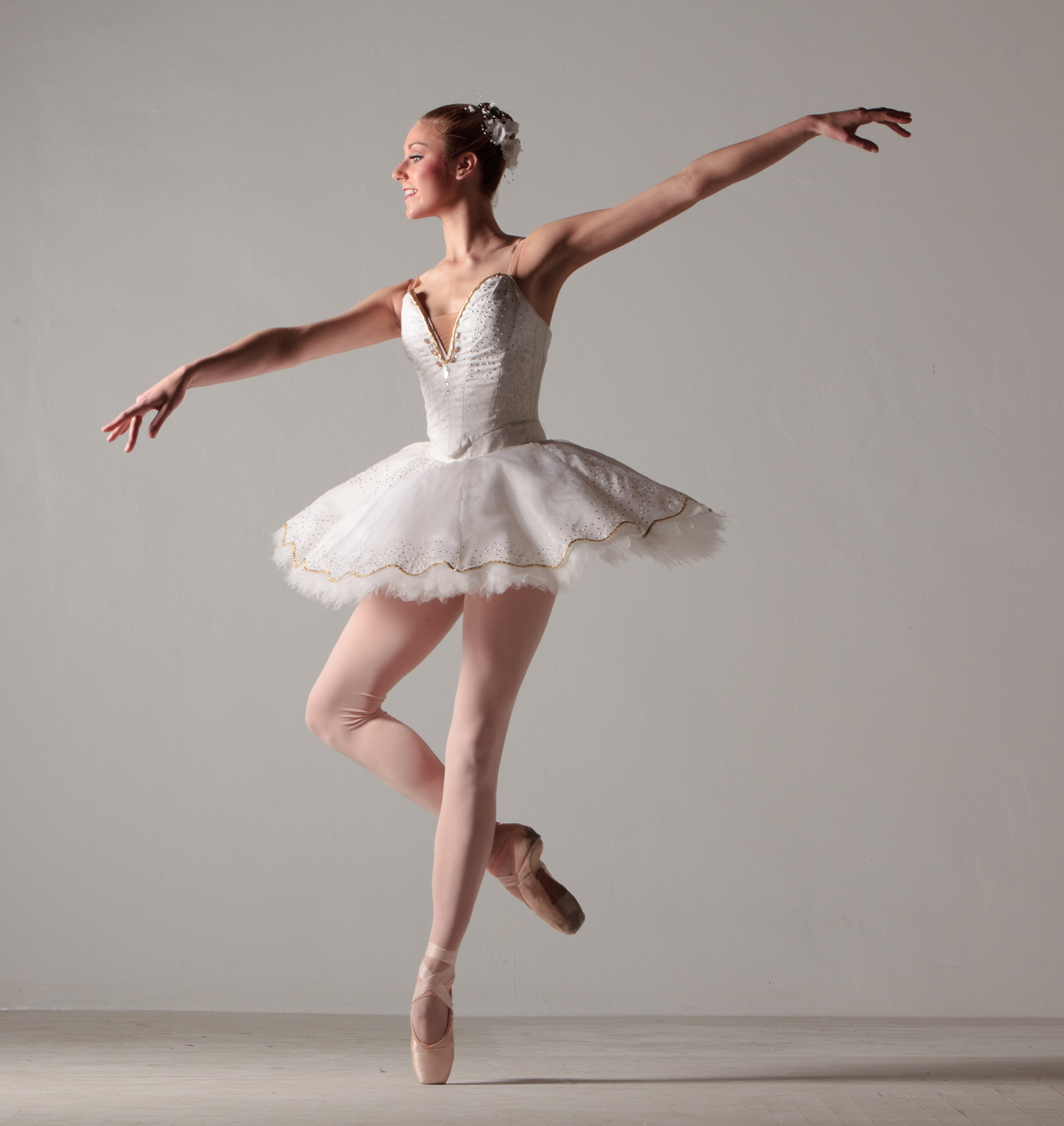}
		\includegraphics[width=0.45\linewidth]{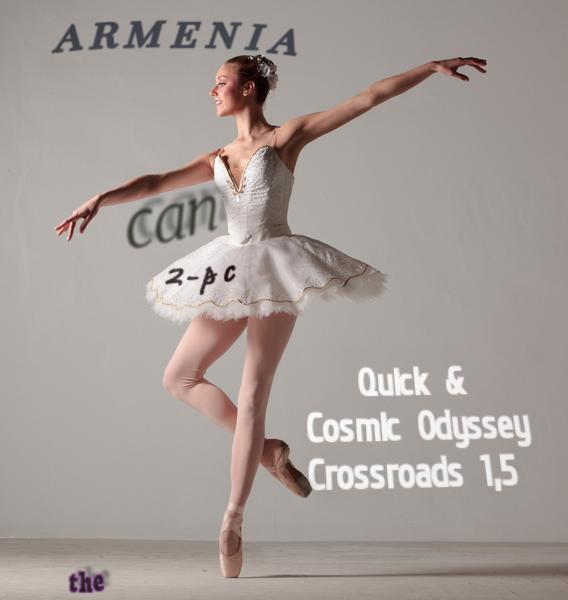}
	}
	\\[0.5ex]
	\subfloat{
		\includegraphics[width=0.45\linewidth]{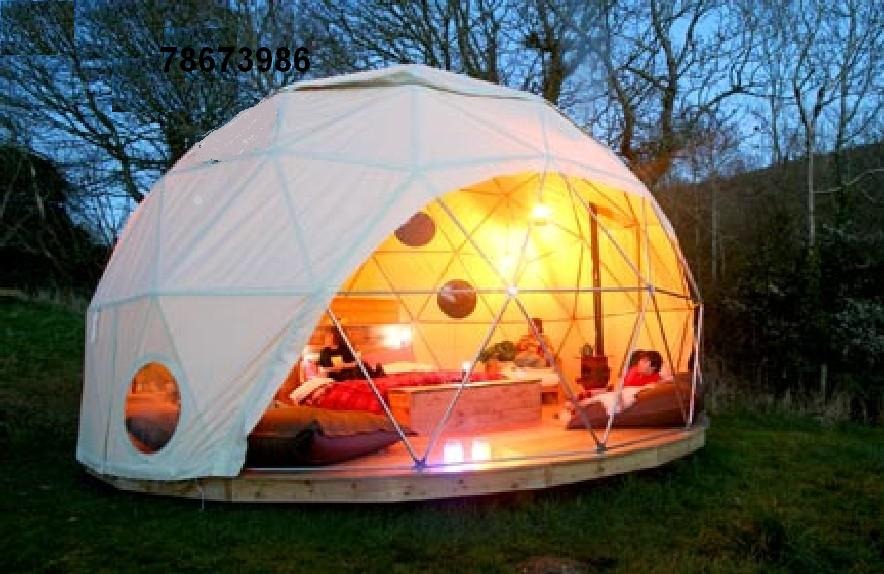}
		\includegraphics[width=0.45\linewidth]{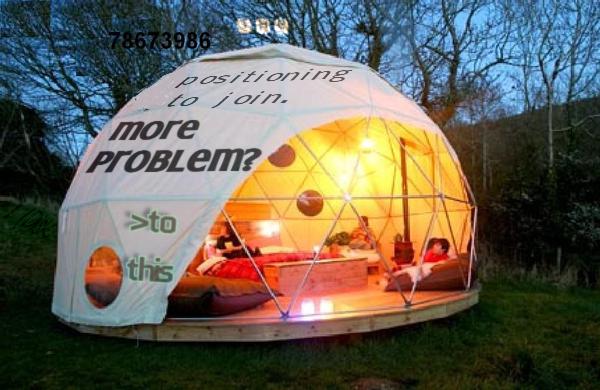}
	}
	\caption{Examples from the synthetic dataset. Left is the background image, right is the synthetic text image.}
	\label{fig:syn_data}
\end{figure}

\noindent
{\bfseries{Synthetic Dataset}} A challenging large-scale synthetic text dataset \cite{gupta2016synthetic} is used to train the model. The dataset includes 800,000 images with nearly 8 million synthetic English word instances. This dataset was initially created for text localization. It, therefore, is a very challenging dataset for training text removal because of various fonts, perspectives and sparsity. In this study, 75\% of the dataset is used as the training set. Examples of this synthetic dataset are given in Fig. \ref{fig:syn_data}. 
\\[1.5ex]
\noindent
{\bfseries{Real-world Datasets}} The generality of the proposed method is evaluated on the test or validation sets of the following three public real scene datasets, which are widely used for scene text detection:

\begin{itemize}
	\item {\bfseries{ICDAR 2013}} \cite{karatzas2013icdar} contains 229 training and 223 testing images with text in English. It is used for both training and evaluation in previous text removal studies \cite{nakamura2017scene,zhang2018ensnet}. In this study, only the test set is used for evaluation for a fair comparison.
	\item {\bfseries{ICDAR 2017 MLT}} \cite{nayef2017icdar2017} is a multi-lingual text dataset, which is composed of 7,200 training images, 1,800 validation images, and 9,000 testing images. Six different scripts are found at arbitrary position in these images. It, therefore, is a very challenging dataset. In this study, the validation set is used for evaluation, since our model is trained on a synthetic dataset, and the test set is only available via special request.
	\item {\bfseries{CTW1500}} \cite{yuliang2017detecting} is real scene curved text dataset. It contains 1,500 images. Each image contains at least one curved text instance. It is a challenging dataset with varied scenes and multi-lingual scripts. In this study, the test set is used for evaluation, which includes 500 images.
\end{itemize}

\noindent
{\bfseries{Mask Generation}} All datasets contain groundtruth bounding boxes with text localization. Masks are generated by setting the values of pixels within the ground truth bounding boxes to one, while all others are set to zero. Note word-level bounding boxes of the synthetic dataset and curve bounding boxes of CTW1500 dataset are used.


\noindent
{\bfseries{Evaluation Metrics}}
In this study, to measure quantitative results, we follow the evaluation method proposed by Nakamura \etal \cite{nakamura2017scene}. In this method, precision, recall and f-score of a baseline text detector is calculated on a text removed dataset. As per Zhang \etal \cite{zhang2018ensnet}, the EAST text detector \cite{zhou2017east} is used, and the results are calculated with two protocols, the DetEval \cite{wolf2014evaluation} and the ICDAR 2013 evaluation \cite{karatzas2013icdar}. Additionally, qualitative results are provided for visual comparison.

\subsection{Comparison with State-of-the-Art}
MTRNet is compared with previous state-of-art methods considering the following four aspects:


\begin{table*}[t!]
	\centering
	\caption{Text Detection Performance on ICDAR 2013 and CTW1500.}
	\begin{tabular}{l|l|c|c|c|c|c|c|c|c|c}
		\hline
		\multirow{3}{*}{Methods}  & \multirow{3}{*}{Resize} & \multicolumn{6}{c|}{ICDAR2013} & \multicolumn{3}{c}{CTW1500}\\\cline{3-11}
		
		&   & \multicolumn{3}{c|}{ICDAR Eval} &  \multicolumn{3}{c|}{DetEval} & \multicolumn{3}{c}{CocoEval}  \\\cline{3-11}
		
		&  & Recall & Precision & f-score & Recall & Precision & f-score & Recall & Precision & f-score \\\hline
		
		Original images	& - &70.10  &81.50  & 75.37 & 70.70 & 81.61 & 75.77 & 68.55  & 75.87  & 72.02\\  
		Original images & $256 \times 256$ &52.91 &80.51  &63.87  & 53.73 & 80.86 & 64.57 & 54.69  & 81.06  & 65.32\\
		Scene text eraser \cite{nakamura2017scene} & - & 22.35 & 30.12 & 25.66 & 34.48 & 60.57 & 43.95 & - &  - & -\\
		Pix2Pix (retrained by \cite{zhang2018ensnet}) & $512 \times 512$ & 10.19 & 69.45 & 17.78 & 10.37 & 69.45 & 18.05 & - & - & -\\
		EnsNet \cite{zhang2018ensnet} & $512 \times 512$ & 5.66 & 73.42 & 10.51 & 5.75 & 73.42 & 10.67 & - & - & -\\
		MTRNet (no mask)  & $256 \times 256$ & 29.11 & 76.05 & 42.11 & 27.83 & 75.85 & 40.73 &  41.20 & 78.51 & 54.04\\
		MTRNet 			& $256 \times 256$ & \bf 0.18 & \bf 16.67 & \bf 0.36 & \bf 0.18 & \bf 16.67 & \bf 0.36 & \bf 0 & \bf 0 & \bf 0 \\\hline
	\end{tabular}
	\label{tab:icdar2013}
\end{table*}
\noindent
{\bfseries{Quantitative}} MTRNet is quantitatively compared with scene text eraser \cite{nakamura2017scene}, Pix2Pix \cite{isola2017image} and EnsNet \cite{zhang2018ensnet}. All these models, unlike MTRNet, are trained by Zhang \etal \cite{zhang2018ensnet} on a dataset which includes images from the training sets of ICDAR 2013 and ICDAR 2017 MLT. While they are compared with MTRNet on the test set of the ICDAR 2013 dataset. Additionally, to further investigate the effectiveness of the mask, MTRNet is also compared with its no mask version on the test set of ICDAR 2013 and the validation set of the CTW1500 datasets. The comparison is shown in Tab. \ref{tab:icdar2013} where the size of the tested images are provided to consider the negative effects of low resolution on text detection.

MTRNet achieved the best performance for both the ICDAR 2013 and CTW1500 datasets. It surpassed the state-of-the-arts with a large margin on the ICDAR 2013 dataset, despite the fact they are trained on a dataset that includes images from ICDAR 2013 and ICDAR 2017 MLT. Moreover, the large margin between performance of MTRNet with and without the mask on both datasets demonstrates that the mask is very beneficial for the text removal performance of MTRNet.

\begin{figure}[!t]
	\centering
	\subfloat{
		\includegraphics[width=0.3\linewidth]{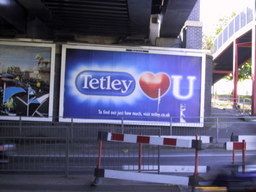}
		\includegraphics[width=0.3\linewidth]{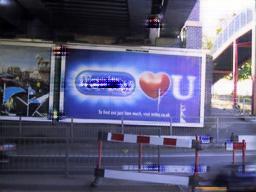}
		\includegraphics[width=0.3\linewidth]{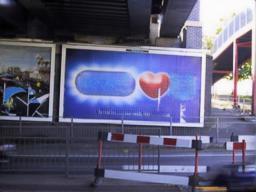}
	}
	\\[0.5ex]
	\subfloat{
		\includegraphics[width=0.3\linewidth]{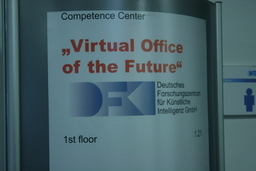}
		\includegraphics[width=0.3\linewidth]{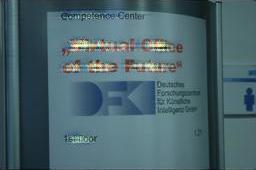}
		\includegraphics[width=0.3\linewidth]{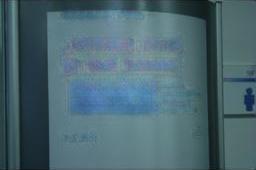}
	}
	\\[0.5ex]
	\subfloat{
		\includegraphics[width=0.3\linewidth]{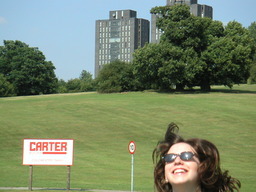}
		\includegraphics[width=0.3\linewidth]{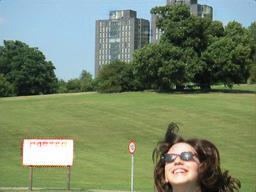}
		\includegraphics[width=0.3\linewidth]{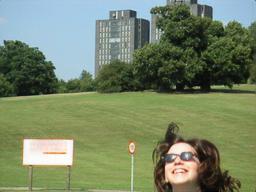}
	}
	\\[0.5ex]
	\subfloat{
		\includegraphics[width=0.3\linewidth]{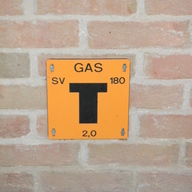}
		\includegraphics[width=0.3\linewidth]{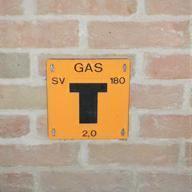}
		\includegraphics[width=0.3\linewidth]{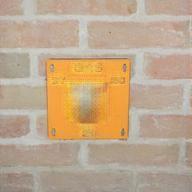}
	}
	\\[0.5ex]
	\subfloat{
		\includegraphics[width=0.3\linewidth]{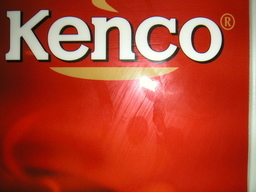}
		\includegraphics[width=0.3\linewidth]{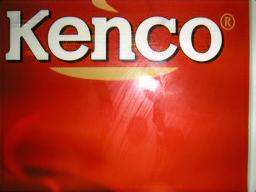}
		\includegraphics[width=0.3\linewidth]{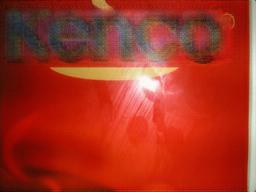}
	}
	\caption{Visual comparison of mask and nomask MTRNet outputs on sample images from ICDAR 2013. In each row from left to right: input, MTRNet nomask output, and MTRNet output.}
	\label{fig:icdar2013}
\end{figure}

\noindent
{\bfseries{Qualitative}} MTRNet is also qualitatively compared with the aforementioned methods on the ICDAR 2013, ICDAR 2017 MLT and CTW1500 datasets. Qualitative results of scene text eraser, Pix2Pix and EnsNet on examples from ICDAR 2013 are given in \cite{nakamura2017scene} and \cite{isola2017image}.  We, therefore, select the same examples from ICDAR 2013 to test MTRNet. According to results shown in \cite{nakamura2017scene} and \cite{isola2017image} and Fig. \ref{fig:icdar2013}, MTRNet shows similar qualitative results to the state-of-the-art, EnsNet, when the size of text is not large, otherwise MTRNet displays acceptable results. The reasons for EnsNet's superior performance on large size text are: 1. EnsNet is trained with ICDAR 2013 and ICDAR 2017 MLT images with large text, while MTRNet is trained on a synthetic dataset where images mainly include small text. 2. EnsNet is trained with special losses, which are beneficial for generating qualitative content, while MTRNet is only trained with adversarial and L1 losses.  MTRNet, therefore, has the potential of reaching the qualitative performance of EnsNet with the same configuration. 

\begin{figure}[!t]
	\centering
	\subfloat{
		\includegraphics[width=0.3\linewidth]{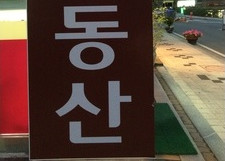}
		\includegraphics[width=0.3\linewidth]{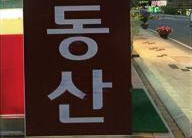}
		\includegraphics[width=0.3\linewidth]{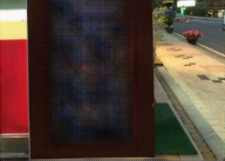}
	}
	\\[0.5ex]
	\subfloat{
		\includegraphics[width=0.3\linewidth]{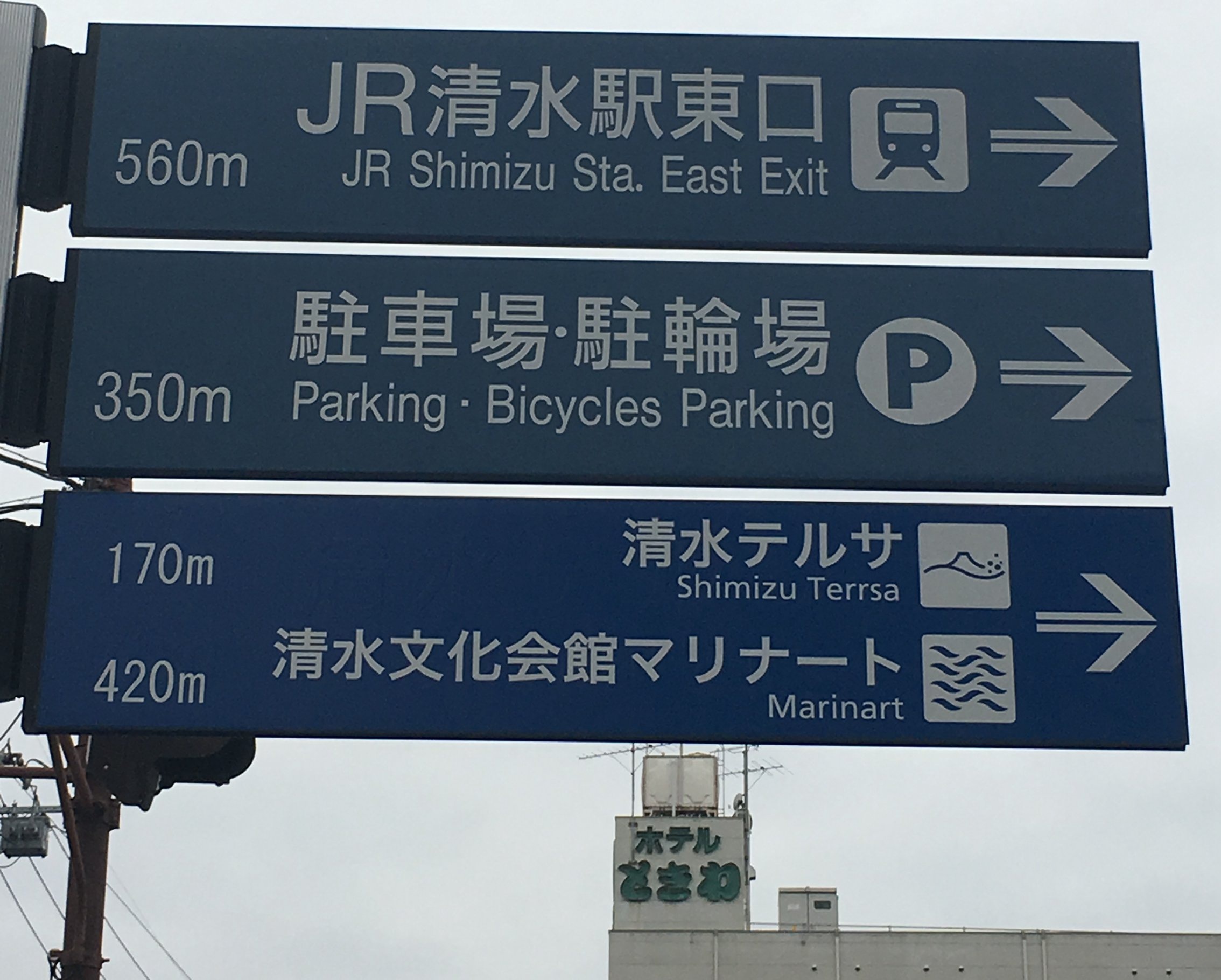}
		\includegraphics[width=0.3\linewidth]{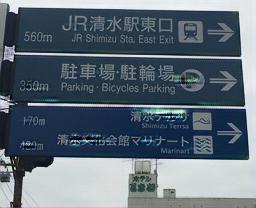}
		\includegraphics[width=0.3\linewidth]{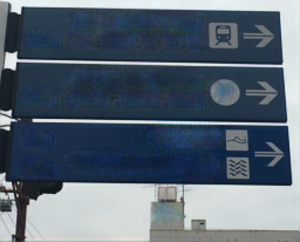}
	}
	\\[0.5ex]
	\subfloat{
		\includegraphics[width=0.3\linewidth]{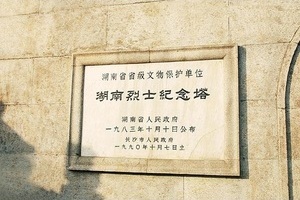}
		\includegraphics[width=0.3\linewidth]{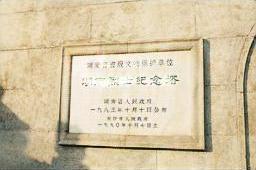}
		\includegraphics[width=0.3\linewidth]{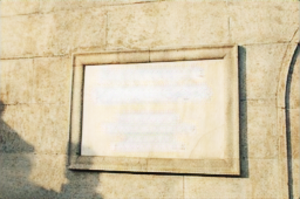}
	}
	\\[0.5ex]
	\subfloat{
		\includegraphics[width=0.3\linewidth]{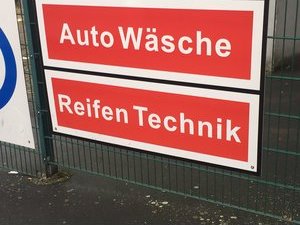}
		\includegraphics[width=0.3\linewidth]{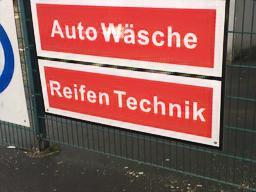}
		\includegraphics[width=0.3\linewidth]{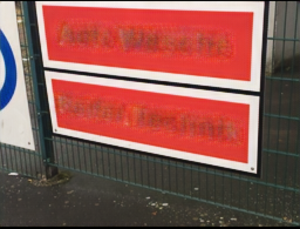}
	}
	\\[0.5ex]
	\subfloat{
		\includegraphics[width=0.3\linewidth]{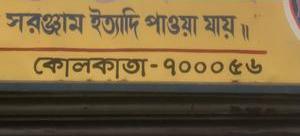}
		\includegraphics[width=0.3\linewidth]{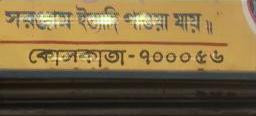}
		\includegraphics[width=0.3\linewidth]{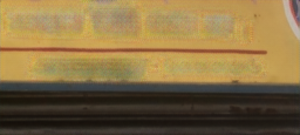}
	}
	\\[0.5ex]
	\subfloat{
		\includegraphics[width=0.3\linewidth]{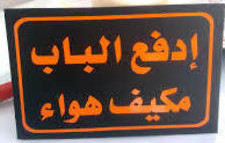}
		\includegraphics[width=0.3\linewidth]{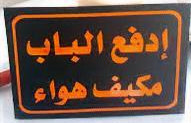}
		\includegraphics[width=0.3\linewidth]{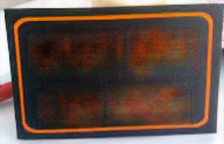}
	}
	
	\caption{Examples of text removal on multi-lingual scripts. Examples from top to down are Korean, Japanese, Chinese, Latin, Bangla and Arabic. In each row from left to right: input, MTRNet nomask output, and MTRNet output.}
	\label{fig:mlt}
\end{figure}

\begin{figure}[!t]
	\centering
	\subfloat{
		\includegraphics[width=0.3\linewidth]{images/curve_text_examples/1011}
		\includegraphics[width=0.3\linewidth]{images/curve_text_examples/1011_rm_pix2pix}
		\includegraphics[width=0.3\linewidth]{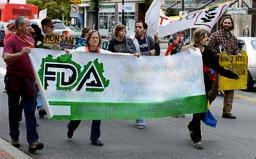}
	}
	\\[0.5ex]
	\subfloat{
		\includegraphics[width=0.3\linewidth]{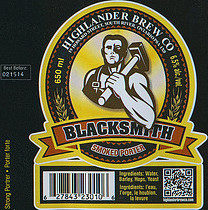}
		\includegraphics[width=0.3\linewidth]{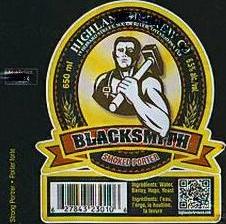}
		\includegraphics[width=0.3\linewidth]{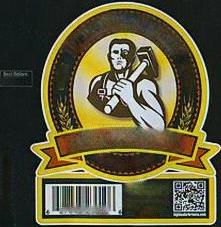}
	}
	\\[0.5ex]
	\subfloat{
		\includegraphics[width=0.3\linewidth]{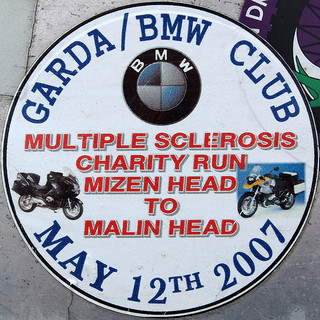}
		\includegraphics[width=0.3\linewidth]{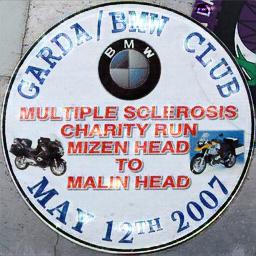}
		\includegraphics[width=0.3\linewidth]{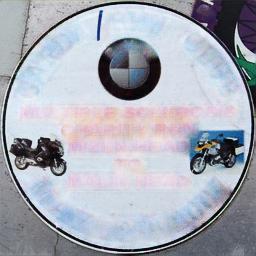}
	}
	\caption{Examples of text removal on curved scripts. In each row from left to right: input, MTRNet nomask, MTRNet.}
	\label{fig:curve}
\end{figure}

Additionally, as shown in Fig.  \ref{fig:mlt} and \ref{fig:curve}, MTRNet shows similar performance on ICDAR 2017 MLT and CTW1500 datasets as well, despite the fact that our synthetic dataset is composed of images with English text in conventional layout and shape. However, without the mask, MTRNet fails in all cases. This suggests that previous state-of-the-art methods \cite{zhang2018ensnet,nakamura2017scene} might be not cable of removing multi-lingual or curved text, as MTRNet without the mask shares a similar network structure to them.



\begin{figure}
	\centering
	\subfloat{
		\includegraphics[width=0.45\linewidth]{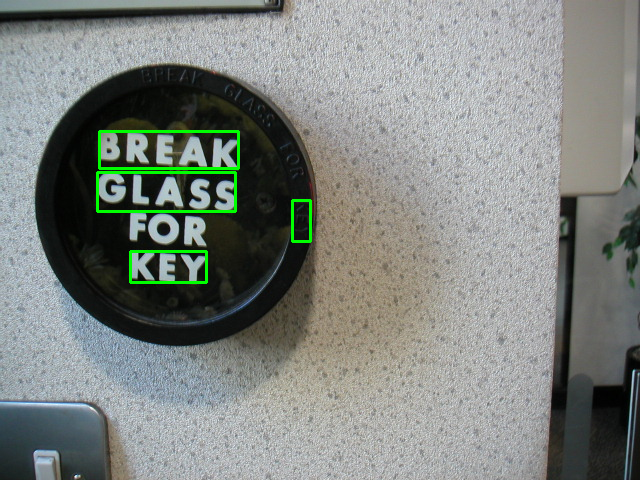}
		\includegraphics[width=0.45\linewidth]{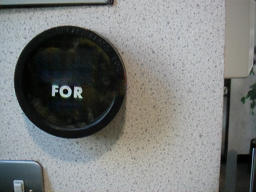}
	}
	\\[0.5ex]
	\subfloat{
		\includegraphics[width=0.45\linewidth]{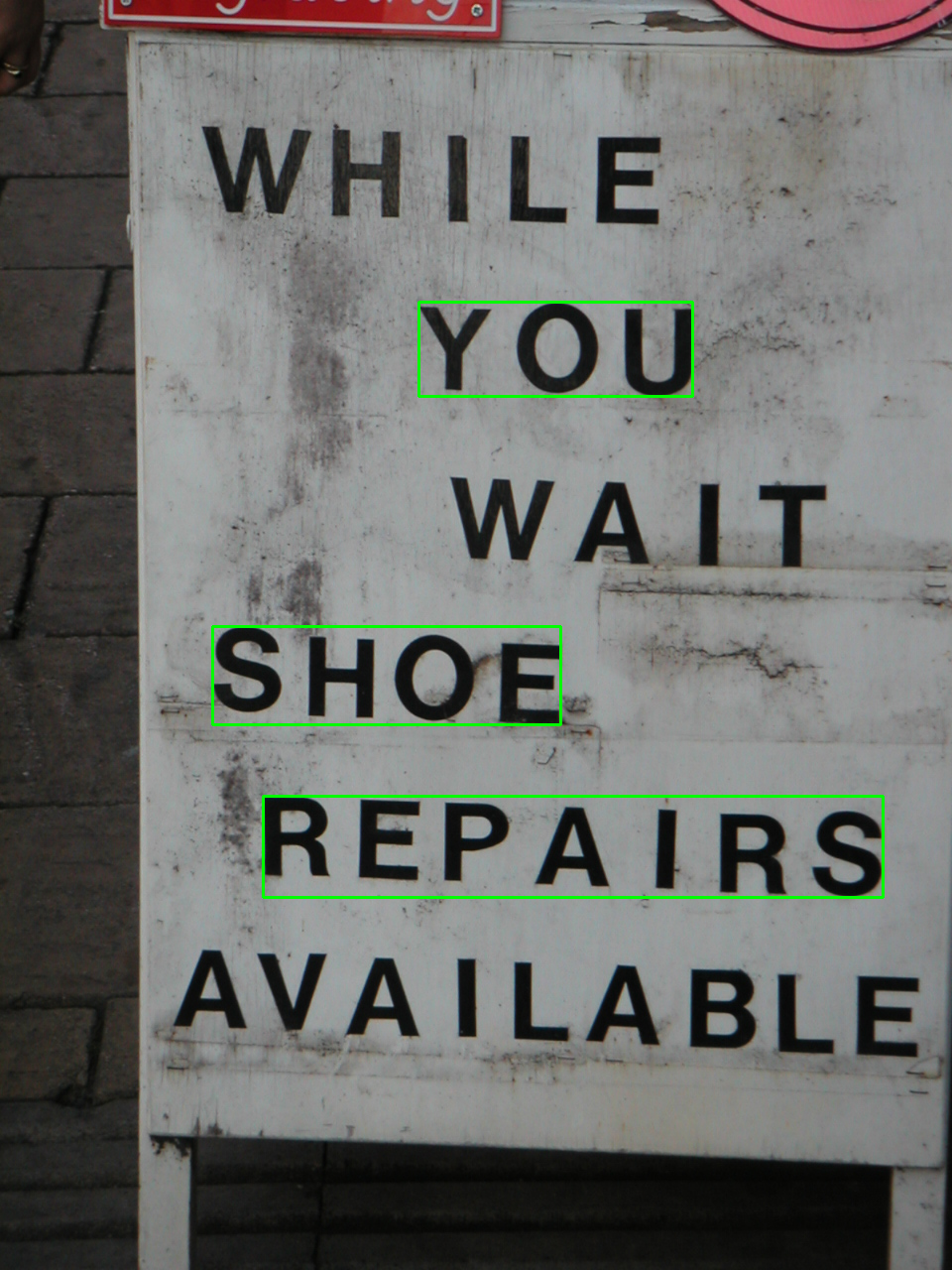}
		\includegraphics[width=0.45\linewidth]{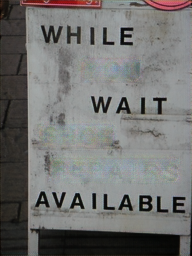}
	}
	\caption{Examples of partial text removal on real scenes. MTRNet is able to partially remove text from a scenes when bounding boxes of selected text are provided. Input images with selected bounding boxes in green are shown in left column, and the partial text removal results are displayed in right column.}
	\label{fig:partial}
\end{figure}

\noindent
{\bfseries{Partial Text Removal}} is not supported by previous state-of-art methods as we discussed in Sec. \ref{sec:intro}. While MTRNet is capable of performing partial text removal as shown in Fig. \ref{fig:partial}, although, during MTRNet training, no special process is applied to gain this partial text removal ability. In training, masks of all text instances in each example are generated. Nevertheless, MTRNet is capable of removing text partially in a scene when masks of selected text are generated at the inference stage.

To test this, masks are generated for randomly selected bounding boxes of text appearing in an image. For example, as shown in Fig. \ref{fig:partial}, masks are generated for texts in green bounding boxes. As a result, MTRNet only removes text in the green bounding boxes, while other text regions remain.

%
\begin{figure}[t!]
	\centering
	\subfloat[Mask]{\includegraphics[width=0.24\textwidth]{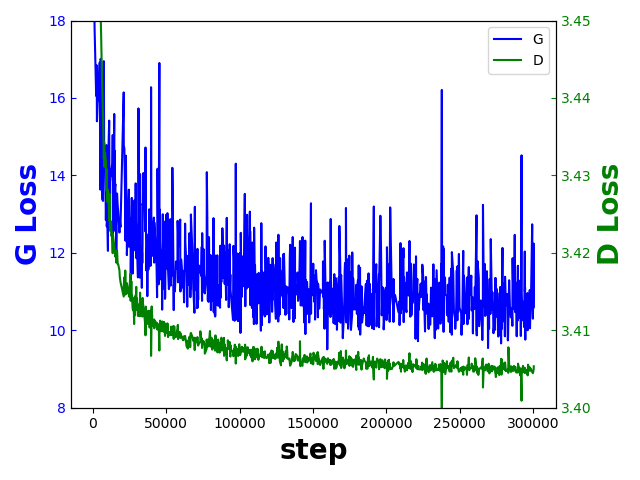}}
	\subfloat[No mask]{\includegraphics[width=0.24\textwidth]{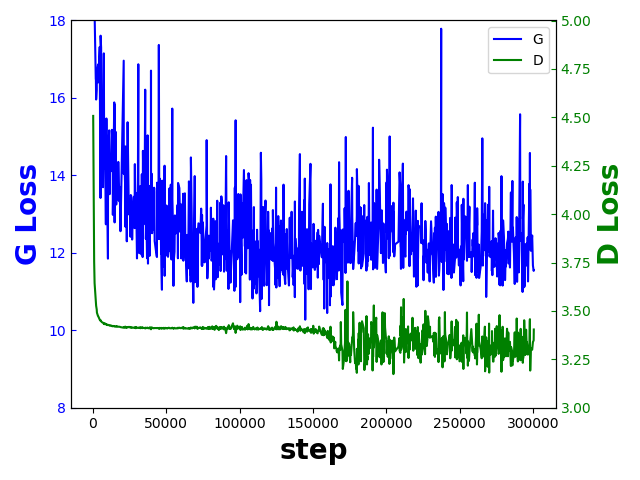}}
	\caption{Training loss for the generator and discriminator when the mask is used (a) and when no mask is used (b).}
	\label{fig:train_curve}
\end{figure}

%
\noindent
{\bfseries{Convergence}} As shown in Fig. \ref{fig:train_curve}, MTRNet has smoother loss curves over time compared to its no mask variant. Without the mask, both the discriminator and generator of MTRNet converge slowly and become very unstable on a large scale dataset with sparse text. The mask frees a model from learning text detection and lets it focus more on inpainting. However, previous state-of-the-art methods have to grasp with text detection functionality. They, therefore, might converge slowly and have an unstable training process on a large-scale challenging datasets.



\section{Conclusion}
\label{sec:con}
In this study, we propose a generic scene text eraser, MTRNet. It addresses the lack of a generic method for text removal from all kinds of scenes with all kinds of text. In this work, MTRNet is trained on a synthetic text detection dataset, and it surpasses previous studies trained on real scene datasets for several benchmark datasets. MTRNet achieves this by appending text masks with the input image prior to passing the input to a cGAN. Text masks not only improve the results of the cGAN, but also make the training of the cGAN on a challenging synthetic dataset more stable and efficient.

\bibliographystyle{ieee}
\bibliography{main}

\end{document}